\documentclass[letterpaper, 10 pt, conference]{ieeeconf}  

\IEEEoverridecommandlockouts                              

\usepackage{graphicx} 
\usepackage{amsmath} 
\usepackage{amssymb}  

\makeatletter
    \let\NAT@parse\undefined
\makeatother
\usepackage[square,numbers,sort&compress]{natbib}
\usepackage{xcolor}
\usepackage{booktabs}
\usepackage{multirow}
\usepackage{bm}
\usepackage{subcaption}
\usepackage{xcolor}
\usepackage{adjustbox}
\usepackage{siunitx}
\usepackage[colorlinks = true,
            linkcolor = black,
            urlcolor  = blue,
            citecolor = magenta,
            bookmarks = true, pagebackref]{hyperref}

\newcommand{\mosaicbgcolor}{white}
\newcommand{\methodname}{PlaceNav}

\title{\LARGE \bf
\methodname: Topological Navigation through Place Recognition
}

\author{Lauri Suomela$^{1}$, Jussi Kalliola, Harry Edelman and Joni-Kristian Kämäräinen
\thanks{ %
$^{1}${\tt lauri.a.suomela@tuni.fi}.
$^{\dagger}$ \href{https://lasuomela.github.io/placenav/}{lasuomela.github.io/placenav/}
}
\thanks{ %
 All authors are with Tampere University, Finland.
This work was supported by Technology Innovation Insitute.}
}

\begin{document}

\makeatletter
\DeclareRobustCommand\onedot{\futurelet\@let@token\@onedot}
\def\@onedot{\ifx\@let@token.\else.\null\fi}

\def\eg{\emph{e.g}\onedot} \def\Eg{\emph{E.g}\onedot}
\def\ie{\emph{i.e}\onedot} \def\Ie{\emph{I.e}\onedot}
\def\cf{\emph{c.f}\onedot} \def\Cf{\emph{C.f}\onedot}
\def\etc{\emph{etc}\onedot} \def\vs{\emph{vs}\onedot}
\def\wrt{w.r.t\onedot} \def\dof{d.o.f\onedot}
\def\etal{\emph{et al}\onedot}

\makeatother


\newcommand{\map}{\mathcal{M}}

\newcommand{\wppolicy}{\pi}

\newcommand{\obs}{I_t}

\newcommand{\zObs}{\mathbf{z}_t}

\newcommand{\node}{I_{s}}

\newcommand{\zNode}{\mathbf{z}_{s}}

\maketitle
\thispagestyle{empty}
\pagestyle{empty}

\begin{abstract}
Recent results suggest that splitting topological navigation into robot-independent and robot-specific components improves navigation performance by enabling the robot-independent part to be trained with data collected by robots of different types.
However, the navigation methods' performance is still limited by the scarcity of suitable training data and they suffer from poor computational scaling.
In this work, we present~\methodname, subdividing the robot-independent part into navigation-specific and generic computer vision components. We utilize visual place recognition for the subgoal selection of the topological navigation pipeline. 
This makes subgoal selection more efficient and enables leveraging large-scale datasets from non-robotics sources, increasing training data availability.
Bayesian filtering, enabled by place recognition, further improves navigation performance by increasing the temporal consistency of subgoals.
Our experimental results verify the design and the new method obtains a \SI{76}{\percent} higher success rate in indoor and \SI{23}{\percent} higher in outdoor navigation tasks with higher computational efficiency.

\end{abstract}

\section{Introduction}
\label{sec:intro}
Autonomous visual navigation is a well-studied problem in the field of robotics~\cite{brooksVisualMapMaking1985a, baumgartnerAutonomousVisionbasedMobile1994}.
One line of research frames navigation in known environments as \textit{topological navigation}~\cite{thrunLearningMetrictopologicalMaps1998, goedemeOmnidirectionalVisionBased2007, filliatInteractiveLearningVisual2008}, meaning purely vision-based navigation between nodes of a topological map that are represented by images.
The advantage of this approach is that it does not require building a geometric reconstruction of the operating environment~\cite{rosinol_kimera_2020} or training environment-specific control policies~\cite{mirowski_learning_2016}.

Topological navigation systems have two parts: \emph{subgoal selection} and a \emph{goal-reaching policy}. First, the subgoal selection module chooses the map node to reach next as a subgoal. Then, the goal-reaching policy produces control commands to take the robot to the selected subgoal.
A popular approach to subgoal selection utilizes \emph{temporal distance prediction}, which means predicting the number of time steps between the robot's current camera observation and the subgoal candidates~\cite{savinov_semi-parametric_2018, savinovEpisodicCuriosityReachability2018, shah_ving_2021, mezghanMemoryAugmentedReinforcementLearning2022, shahGNMGeneralNavigation2023, shahViNTLargeScaleMultiTask2023}. These learning-based models are trained with offline datasets of robot trajectories.

Previous works have demonstrated impressive real-world navigation~\cite{shah_ving_2021, shahGNMGeneralNavigation2023, shahViNTLargeScaleMultiTask2023}, but the temporal distance prediction
approach has two significant shortcomings.
First, the distance has to be estimated individually for each subgoal candidate, incurring complexity that scales at $\mathcal{O}(n)$ with the number of candidate images. This
requires heuristics to limit the candidates and constrains the methods available for ensuring subgoal temporal consistency.
Second, the fact that temporal distance prediction requires training data that originates from robots, actual or simulated, introduces an unnecessary data bottleneck. High-quality robotics datasets are very scarce compared to general web-scale data, and models trained with simulated data suffer from generalization issues~\cite{kar_meta-sim_2019}.

\begin{figure}
    \centering
    \includegraphics[width=0.86\linewidth]{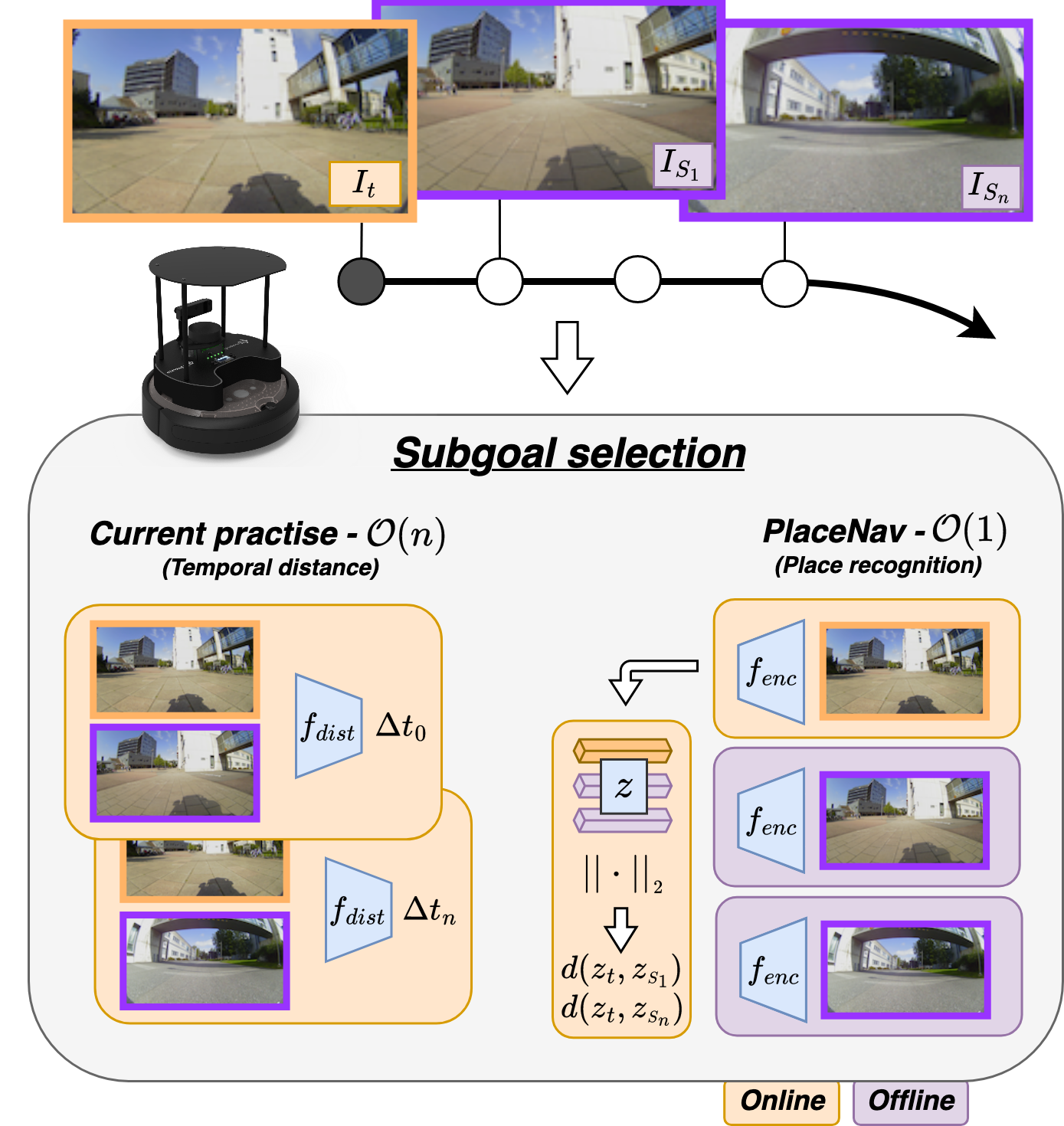}
    \caption{Visual place recognition finds which map image $I_s$ was captured closest to the robot's observation $I_t$ by efficient matching of image embeddings $\bm{z}$.}
    \label{fig:opener}
\end{figure}

We claim that subgoal selection is not a unique problem but an instance of the broader concept of image retrieval. To address this, we present\textbf{\emph{~\methodname}}, which frames the selection as a \emph{place recognition}~\cite{masone_survey_2021} task. This design provides three advantages.
First, the large-scale and high-diversity datasets available for training place recognition models enhance subgoal selection robustness against changes in viewpoint and appearance.
Second, subgoal selection is performed by a fast nearest-neighbor search over image embeddings.
This, as illustrated in Fig.~\ref{fig:opener}, provides superior scalability and removes the need for heuristics.
Finally, place recognition readily integrates with methods for ensuring temporal consistency.

In summary, our contributions are

\begin{itemize}

    \item A navigation approach that decouples training of subgoal selection models from robotics datasets by treating the selection as a generic place recognition task. 

    \item Integration of learning-based subgoal selection with a Bayesian filter that improves temporal consistency.

    \item Demonstrations with real robots that experimentally validate our design.

\end{itemize}

Code and videos are available on the project page$^{\dagger}$.
	
\section{Related work}
\label{sec:related}
\noindent\textbf{Vision-based topological navigation.}
In a recent trend, topological maps have been used to divide long-horizon tasks into short-horizon segments suitable for learned goal-reaching policies~\cite{savinov_semi-parametric_2018, singh_chaplot_neural_2020}. An essential part of such a hierarchical structure is choosing which subgoal to reach next. One approach is to learn to predict the reachability between two images from simulated rollouts~\cite{meng_scaling_2020}. Savinov~\etal~\cite{savinov_semi-parametric_2018} proposed to use the \emph{temporal distance}, or the number of time steps $\Delta t$, between the current observation and a subgoal as a proxy for reachability. Its key strength is that it can be learned from offline data. 
While alternative approaches exist~\cite{eysenbach_search_2019}, temporal distance is popular and has been adopted in several recent works~\cite{savinovEpisodicCuriosityReachability2018, shah_ving_2021, mezghanMemoryAugmentedReinforcementLearning2022, shahGNMGeneralNavigation2023, shahViNTLargeScaleMultiTask2023}. 
However, the diversity and size of datasets suitable for temporal distance learning are modest. ~RECON~\cite{shahRapidExplorationOpenWorld2022} with \SI{25}{\hour}, SACSoN~\cite{hirose_sacson_2024} with \SI{75}{\hour}, and TartanDrive~\cite{triestTartanDriveLargeScaleDataset2022} with \SI{5}{\hour} of navigation trajectories from single location each are notable examples.
Furthermore, because of the model architectures utilized, the computational complexity of temporal distance prediction scales at O(n) with the number of subgoal candidates considered.

\textbf{Place recognition.}
Place recognition involves recognizing places from images, often framed as image retrieval across images captured from different viewpoints and varying appearances~\cite{masone_survey_2021}. It naturally integrates with topological navigation, as subgoal selection can be viewed as an image retrieval problem.
Traditional methods for place recognition rely on aggregating handcrafted local features~\cite{sivic_video_2003, jegou_aggregating_2010, torii_247_2015}, but newer methods utilize deep learning to extract embeddings that can be compared efficiently using nearest-neighbor search~\cite{arandjelovic_netvlad_2018, revaud_learning_2019, berton_rethinking_2022}. The methods are trained to produce embeddings that are similar for images from the same place and dissimilar for images from different places, typically by classification loss~\cite{berton_rethinking_2022} or ranking losses such as contrastive~\cite{radenovic_fine-tuning_2019}, triplet~\cite{arandjelovic_netvlad_2018} or listwise~\cite{revaud_learning_2019} loss.

\textbf{Temporal consistency.}
While subgoal temporal consistency has been studied in non-learned topological navigation literature~\cite{goedemeOmnidirectionalVisionBased2007}, it has received limited attention in the context of learning-based methods.
A robot moves along a route continually, so the transitions between the subgoals should be smooth. As a heuristic solution, SPTM~\cite{savinov_semi-parametric_2018} and GNM~\cite{shahGNMGeneralNavigation2023} adopted the approach of only considering subgoals within a sliding window centered on the previous subgoal. Meng~\etal~\cite{meng_scaling_2020} utilize a similar approach but resort to global search when the window deviates from the robot's actual location.

In place recognition literature, the topic has received more attention. Early approaches \cite{milford_seqslam_2012, hansen_visual_2014, pepperell_all-environment_2014, naseer_robust_2018} utilized feature similarity matrices to find the best-matching image sequences. A newer line of work~\cite{gawel_x-view_2018, latif_addressing_2018, garg_seqnet_2021} considers descriptors that represent sequences instead of individual images. 
As an alternative, Xu~\etal~\cite{xu_probabilistic_2021, xuProbabilisticAppearanceInvariantTopometric2021} added a Bayesian filter to the matching process. 
In this work, we show that learning-based topological navigation also benefits from such methods.

\section{System overview}
\label{sec:system}
In this section, we describe~\emph{\methodname}, our proposed navigation pipeline. First, we discuss the basic components and definitions of topological navigation. Then, we elaborate on our contributions related to subgoal selection via place recognition and subgoal temporal consistency.

\subsection{Topological navigation fundamentals}

Autonomous navigation using topological maps generally consists of two stages. Before navigation, an operator has to perform a manual 'reference run' to capture the desired route. The robot saves images along the route that compose the topological map $\map$ for navigation.

During navigation, the robot-agnostic topological navigation algorithm is combined with a robot-specific controller.
At each inference step $t$, the current robot observation $\obs$ is compared to the different subgoal candidate images $\node \in \map$ at nodes $~s=[0,1,\ldots,S]$ of the topological map. One of the nodes is selected as the next subgoal, and an image-based goal-reaching policy produces the motion plan to reach it.
In this work, we experiment with the different subgoal selection methods and adopt the waypoint estimation approach proposed by Shah~\etal~\cite{shahGNMGeneralNavigation2023} as the goal-reaching policy. 
This approach defines the motion plan as a sequence of $\tau$ waypoints $\{p_i, \psi_i\}_i, ~i=[0, 1, \ldots, \tau]$ that guide the robot to the subgoal.
The waypoints, defined as metric coordinates $p_i$ and heading angle $\psi_i$ in the robot's local coordinate frame, are tracked by a robot-specific controller.

\subsection{Subgoal selection via place recognition}
\label{sec:subgoal}

\methodname~introduces the following modifications to the subgoal selection procedure. 
Instead of computing temporal distances $\Delta t$ between each observation and subgoal candidate pair, we use a place recognition model $f_{enc}$ to process the observation and map images separately. The model produces image embeddings $\zObs$ and $\zNode$ that can be compared by Euclidean distance, enabling efficient subgoal search. Figure~\ref{fig:opener} visualizes the concept.

\vspace{0.1cm}
\textbf{Training data availability.}
The temporal distance prediction models are trained to predict the $\Delta t$ between two image frames sampled from a robot-driven trajectory. This limits the amount and diversity of potential training data.
Place recognition methods can be trained with data from more generic sources. Training utilizes images of different places, preferably captured at various points in time, from different viewpoints, and under different environmental conditions. The images' rough position and orientation information provide annotations. Google StreetView images, for example, are well-suited as training data. The sizes of place recognition datasets are in the order of millions of images~\cite{carlevaris-biancoUniversityMichiganNorth2015, chen_deep_2017, warburg_mapillary_2020}, the SF-XL~\cite{berton_rethinking_2022} alone consisting of 41M images.

\vspace{0.1cm}
\textbf{Computational complexity.}
The computational complexity of temporal distance prediction scales linearly with the number of subgoal candidates considered at each inference step. Because of this, the number of subgoal candidates must be limited, which is commonly implemented as a \emph{sliding window} over the map nodes. The window is centered on the subgoal from the previous step, and only the nodes within the window are considered potential subgoals.

Place recognition enables computation and storage of the descriptors for the map images offline before robot operation. Thus, the inference computational complexity of subgoal selection is decoupled from the number of subgoal candidates being considered. The descriptors can be matched in milliseconds by nearest neighbor search~\cite{johnson_billion-scale_2021}. Consequently, heuristics to limit the number of subgoal candidates are not needed from the perspective of computational budget.

\subsection{Temporal consistency}
\label{sec:consistency}

Limiting the number of subgoal candidates also enhances temporal coherence between inference steps, preventing erratic subgoal selection~\eg~due to visually similar content. A sliding window over the map achieves this to some extent. However, the window may drift away from the robot's location, making correct subgoal selection impossible. Bayesian filtering, enabled by efficient matching of image embeddings, 
is an alternative strategy for enforcing temporal consistency.

\begin{figure}
    \centering
    \includegraphics[width=0.9\linewidth]{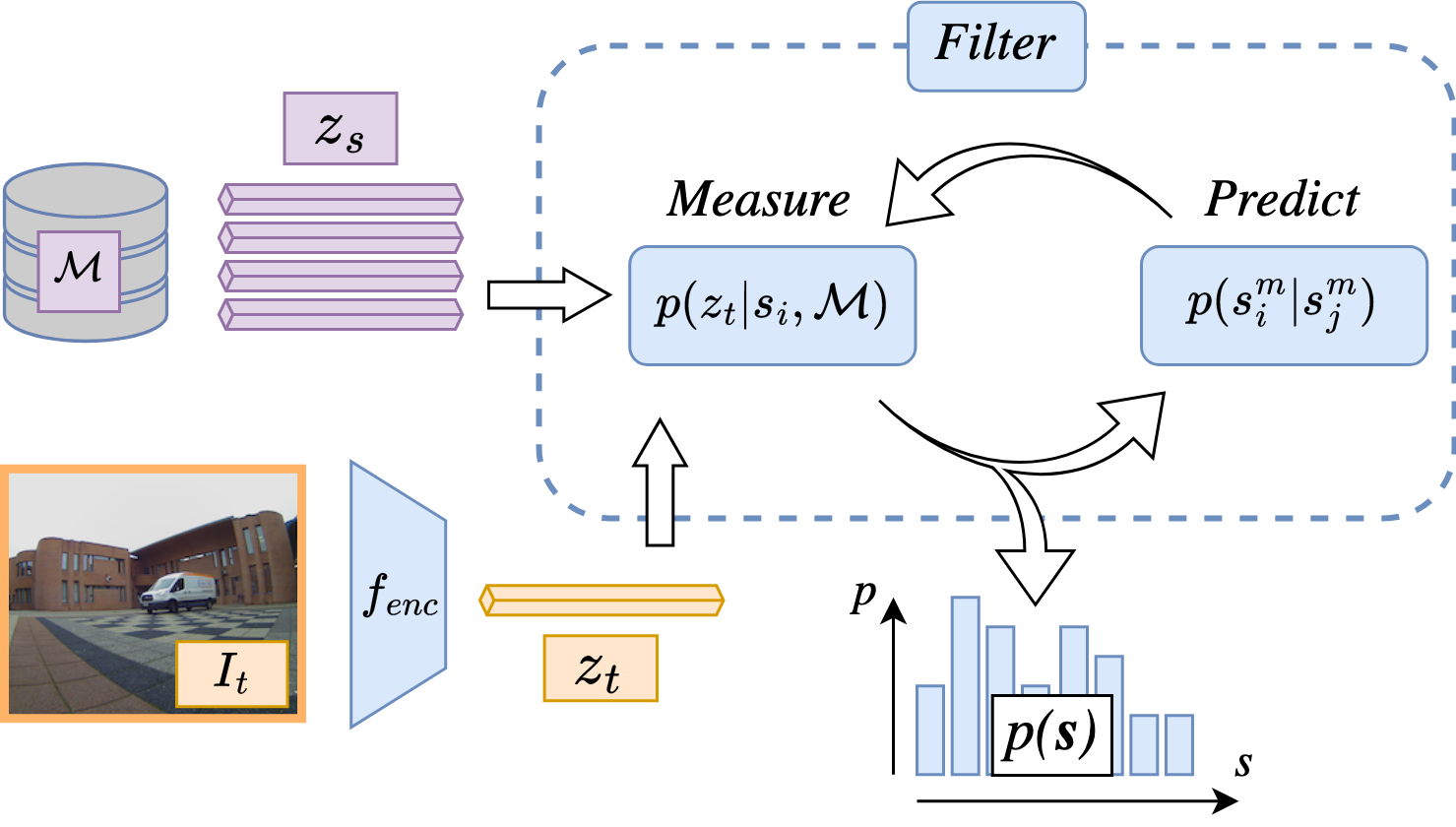}
    \caption{The discrete Bayesian filter alternates place recognition measurement and motion model prediction. $p(\bm{s})$, the posterior belief, determines the best matching node.
    }
    \label{fig:topo-filter}
\end{figure}

Xu~\etal~\cite{xu_probabilistic_2021} propose one such approach for use with place recognition methods, which we adapt for our problem. The idea, illustrated in Fig.~\ref{fig:topo-filter}, is to formulate place recognition as the measurement step of a discrete Bayesian state estimator. This filter maintains a belief of the robot's location over the map nodes by recursively updating its posterior distribution. We present the key equations here for the sake of completeness but refer the reader to the original paper for details.

Given an initial posterior belief distribution, a motion model propagates the belief into the future in a prediction step. If the robot's local movement (\ie~odometry) is not being tracked, as is the case with our system, the motion model is simply 
\begin{equation}\label{eq:motionmodel}
    p(s_{i} | s_{j}) \propto 
    \begin{cases}
    1 & w_l \leq i - j \leq w_u \\
    0 & \text{otherwise}
    \end{cases} \enspace .
\end{equation}
From each node, the robot has an equal probability of moving up to $w_u$ steps toward the goal, staying put, or moving $w_l$ steps back. Other transitions have zero probability.

The prediction step is followed by a measurement step, where a place recognition query between the observation and the map nodes produces a measurement belief by the measurement function
\begin{equation}\label{eq:measurementmodel}
    p( \zObs | s, \map) \propto g(\zObs, s, \map) = \exp(-\lambda_1 \|\zObs - \zNode \|_2) \enspace,
\end{equation}
where $\zObs$ is the observation embedding, $\zNode$ is a map node embedding at the state $s$ being considered, and $\mathcal{M}$ is the map. $\lambda_1$ scales the effect of each measurement on the posterior. Its value is automatically determined at the beginning of each navigation session as proposed by Xu~\etal~\cite{xu_probabilistic_2021}.

The measurement belief is multiplied by the belief from the prediction step to acquire the posterior belief distribution $p(s)$. The map node with the highest posterior probability is considered the closest to the latest observation.
This filter significantly improves the stability of subgoal selection. Unlike the sliding window approach, the filter maintains full posterior belief over all map nodes, so it cannot get lost. It can solve the 'kidnapped robot' problem~\cite{thrun_robust_2001}, whereas the sliding window requires the start node of the robot to be specified manually.

\subsection{Implementation details}

\textbf{Architecture \& Training.}
The Shah~\etal~\cite{shahGNMGeneralNavigation2023} waypoint estimation model was chosen as the goal-reaching policy. We do not retrain the model and use the weights provided by the authors, trained with $85\times 64$ images.

For the place recognition part of the~\methodname, we use a CosPlace network~\cite{berton_rethinking_2022} because of its high performance and simple training process. The model architecture comprises a convolutional encoder and a generalized mean pooling (GeM) layer. The model is trained via classification loss, enabled by dividing the training data into distinct spatial groups.
As the original CosPlace model was trained with high-resolution images ($512\times 512$), the training checkpoints provided by the authors do not work well with the $85\times 64$ images we need to use for comparison with 
the baseline temporal distance model from Shah~\etal~\cite{shahGNMGeneralNavigation2023}.
%
For our experiments, we train a CosPlace model from scratch using an EfficientNet-B0~\cite{tanEfficientNetRethinkingModel2019} backbone and the 41.2M images of the San Francisco eXtra Large (SF-XL) dataset~\cite{berton_rethinking_2022}, resized to $85\times 85$ during training. The model was configured to extract 512-dimensional descriptors. Otherwise, we followed the training procedure outlined in~\cite{berton_rethinking_2022}. We will refer to this low-resolution model as \emph{CosPlace-LR}.


\vspace{0.2cm}
\textbf{Deployment.}
During inference, the robot uses place recognition to identify the map node that best matches the current observation. The next node along the route after the best-matching node is selected as the subgoal.
We implemented two distinct temporal consistency methods in the subgoal selection. The first is the sliding window used in prior works~\cite{savinov_semi-parametric_2018, meng_scaling_2020, shahGNMGeneralNavigation2023}, and the second is our implementation of the discrete Bayesian filter proposed by Xu~\etal~\cite{xu_probabilistic_2021}.
At the beginning of a route, the sliding window is initialized to the first node. The initial belief distribution of the discrete filter is acquired from the first place recognition query, meaning that it operates in a 'kidnapped robot' mode. We set the discrete filter motion model transition boundaries to $w_u=2$ and $w_l=-1$ based on calibration experiments.

After choosing the subgoal to reach next, the goal-reaching policy predicts a series of 5 waypoints $\{p_i, \psi_i\}$. The robot follows these waypoints in an open-loop fashion until the next prediction using the robot's low-level controller. The prediction loop of subgoal selection and waypoint prediction runs at \SI{5}{\hertz}. The place recognition and waypoint estimation models receive $85\times 64$ resolution images as input.

\section{Experimental setup}
\label{sec:experiments}

\begin{figure}
    \centering
    \begin{adjustbox}{width=0.65\linewidth, bgcolor=\mosaicbgcolor} %
    \includegraphics[width=\linewidth]{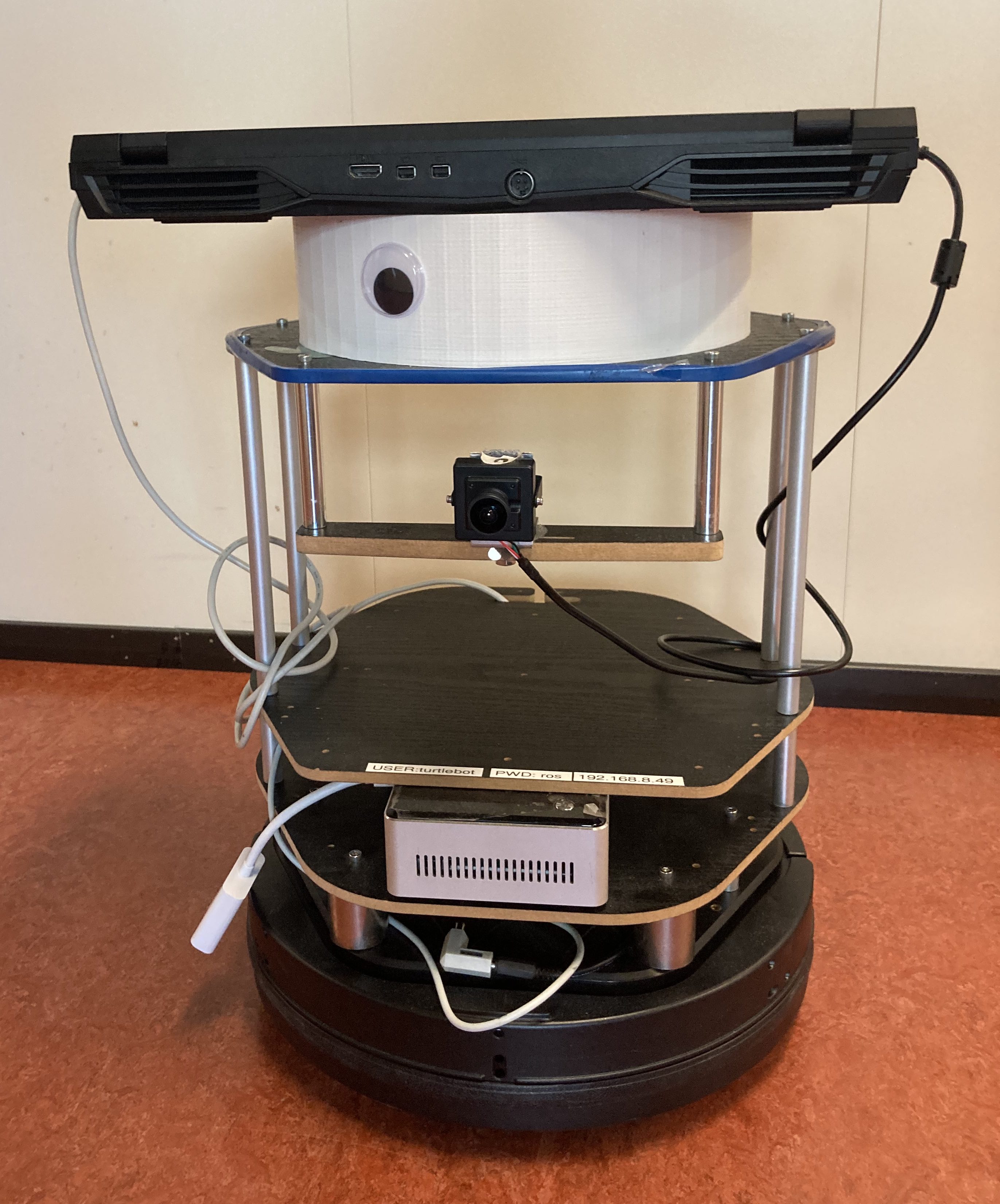}
    
    \hspace{0.15cm}
    
    \includegraphics[trim=0 55 0 55,clip,width=\linewidth]{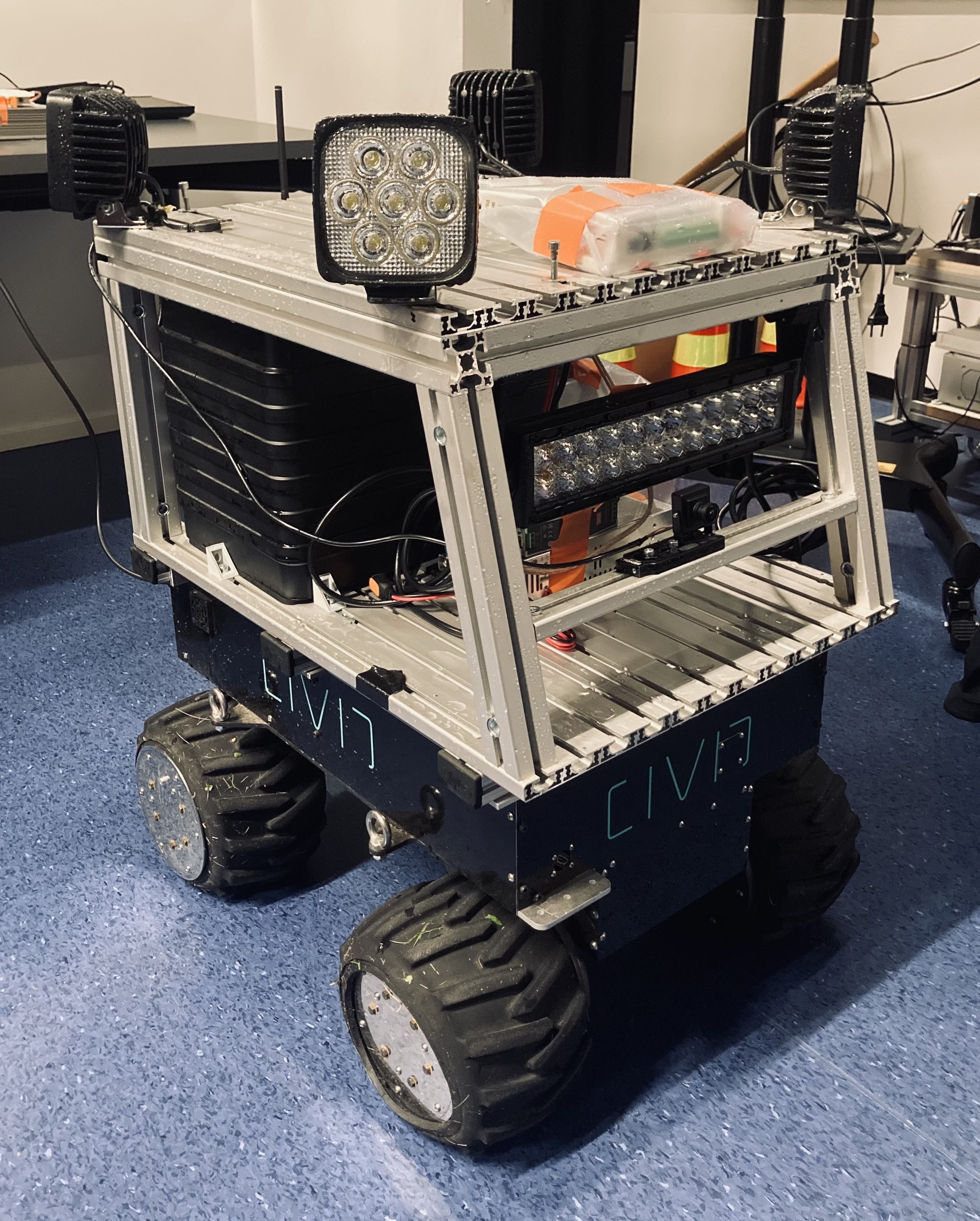}
    \end{adjustbox}
    \caption{The robots: A Turtlebot2 (left) and a Robotnik Summit XL Steel (right) }
    \label{fig:robots}
\end{figure}

We performed navigation experiments with real robots in diverse environments to enable an informed comparison of
\methodname~%
and the prior work.
We conducted 360 repetitions of different routes with different robots, subgoal selection methods, and temporal consistency approaches, adding up to a total of \SI{19}{\km} of navigation testing.
We also evaluated the subgoal selection methods offline using a place recognition benchmark.
With our experiments, we aim to answer the following research questions:

\begin{itemize}

    \item \textbf{Q1.} Does subgoal selection require models trained with data originating from robots, or could more efficient place recognition models replace them?
    
    \item \textbf{Q2.} How robust are temporal distance prediction models to viewpoint and appearance changes?

    \item \textbf{Q3.} How does subgoal temporal consistency affect navigation performance?
    
\end{itemize}

\subsection{Robot navigation experiments}

\textbf{The robots.}
We experimented with two different robots, a Turtlebot2, and a Robotnik Summit XL Steel, shown in Fig.~\ref{fig:robots}. Turtlebot is a small research platform intended for indoor use. Summit is a 4-wheeled skid-steer ground vehicle with a weight of \SI{90}{\kilo \g}, load capacity of \SI{130}{\kilo \g}, and a maximum velocity of \SI{3}{\meter/\second}. Both robots carry a front-facing \SI{175}{\degree} field-of-view fish-eye camera, which is used for navigation, and a laptop computer that runs the navigation algorithms. The laptop has a Nvidia Geforce GTX 1070 GPU and an Intel i7-7700 CPU. The deep learning components of the navigation stack run on the GPU.

\vspace{0.2cm}
\textbf{Baseline.}
We used the temporal distance prediction from GNM~\cite{shahGNMGeneralNavigation2023} by Shah~\etal~as the baseline. GNM was utilized for waypoint prediction in all experiments and we only experiment with the subgoal selection.
We use model weights provided by the authors, obtained by training the model with $85\times64$ images worth 54 hours of navigation. The inference loop runs at \SI{5}{\hertz}. At each step, the node with the smallest temporal distance $\Delta t$ above 3 is selected as the subgoal.

As the Bayesian filter requires the distance between the current observation and all the map nodes at each step, using it with the GNM is not computationally feasible. Thus, with GNM, we only utilize the sliding window.

\begin{figure}
    \centering

    \begin{adjustbox}{width=\linewidth, bgcolor=\mosaicbgcolor, varwidth=10cm} 
    \begin{subfigure}{0.32\linewidth}
        \includegraphics[width=\linewidth]{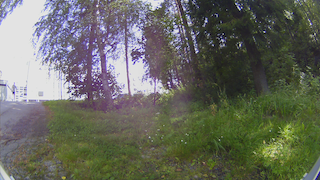}
    \end{subfigure}
    \begin{subfigure}{0.32\linewidth}
        \includegraphics[width=\linewidth]{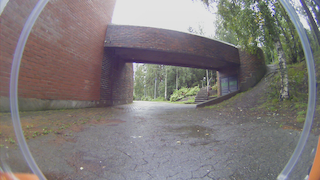}
    \end{subfigure}
    \begin{subfigure}{0.32\linewidth}
        \includegraphics[width=\linewidth]{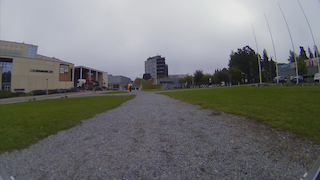}
    \end{subfigure}

    \vspace{0.1cm}
    
    \begin{subfigure}{0.32\linewidth}
        \includegraphics[width=\linewidth]{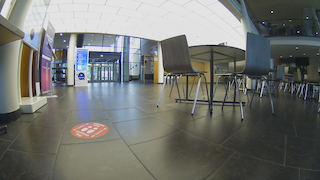}
    \end{subfigure}
    \begin{subfigure}{0.32\linewidth}
        \includegraphics[width=\linewidth]{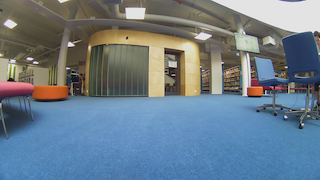}
    \end{subfigure}
    \begin{subfigure}{0.32\linewidth}
        \includegraphics[width=\linewidth]{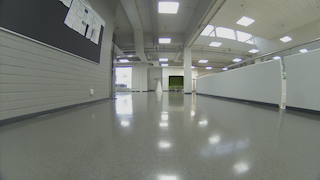}
    \end{subfigure}
    \end{adjustbox}
    
    \caption{Examples along the test routes. The top row is from outdoor tests, bottom row is from indoors.}
    \label{fig:ref-routes}
\end{figure}

\vspace{0.2cm}
\textbf{Indoor navigation.}
The indoor experiments were conducted with the Turtlebot2 robot. We tested the methods along 20 different routes, performing 3 repetitions of each route with each method. Fig.~\ref{fig:ref-routes} shows examples along the routes. Routes where all the methods fail were excluded from the quantitative analysis. The experiments took place at a university campus, with buildings from various eras with diverse appearances. 

The lengths of the test routes were uniformly distributed in the range from \SI{15}{\meter} to \SI{35}{\meter}, and they contained various amounts of difficult content, such as turning maneuvers and passing narrow gaps. We chose routes that do not contain content that would cause navigation failures because of errors in waypoint estimation. Such content,~\eg~very narrow gaps, and turning maneuvers in open areas with few salient features can cause the robot to veer off course even though the subgoal selection algorithm is not failing.

\vspace{0.2cm}
\textbf{Outdoor navigation.}
The outdoor experiments were conducted with the Summit XL robot. We experimented on 20 test routes in an urban environment, ranging from \SI{50}{\meter} to \SI{150}{\meter} in length. Like the indoor experiments, each test route was repeated 3 times with each method.

In indoor calibration tests before the actual experiments, correct subgoal selection led to accurate waypoint estimation. In outdoor tests, we observed more variability. This is likely due to increased environmental noise and larger appearance changes between the reference and test runs. Sometimes changes in ambient illumination led to complete waypoint estimation failure despite a good subgoal choice. In this work we are interested in the performance of subgoal selection, not the goal-reaching policy.
Therefore, in the experiments, we maintained the difference between test and reference run illuminances below a threshold of \SI{10000}{\lux}.

\vspace{0.2cm}
\textbf{Evaluation criteria.}
We follow the evaluation guidelines for image-goal navigation in \cite{anderson_evaluation_2018} and measure the navigation performance by the \emph{success rate} (SR). Success rate describes the ratio of successful and failed repetitions of a test route.

Repetition is considered successful when the robot reaches a pose where its camera view corresponds to the final node of the topological map. The subgoal selection must also localize to this node, triggering a navigation stop signal.
We do not use distance-based success criteria, which are more common but less aligned with the robot's goal determination. While last-mile metric navigation could enhance goal-reaching precision, as suggested by Wasserman~\etal~\cite{wassermanLastMileEmbodiedVisual2023}, it is unnecessary for the scope of this work. Repetition is considered a failure when the robot takes actions that prevent it from reaching the goal,~\ie~sending the 'reached goal' signal without the goal view in sight, risking immediate collisions, or deviating significantly from the test route without recovery prospects.

\subsection{Offline evaluation}

To answer \textbf{Q2.} and enable a reproducible comparison of temporal distance prediction and place recognition, we tested GNM on standard place recognition benchmarks that contain appearance and viewpoint changes.
The VPR-Benchmark by Bertoli~\etal~\cite{bertonDeepVisualGeoLocalization2022} was used to facilitate the experiments. We modified the benchmark to enable the evaluation of GNM that simultaneously takes the query and reference images as inputs. A subset of the VPR-Benchmark datasets where the size of the test database is small enough that GNM can process it in a reasonable time was picked for evaluation. We compared the performance of our low-resolution CosPlace-LR model and
GNM temporal distance.
The test images were resized to $85\times64$. For reference, we additionally evaluate the standard CosPlace model with full-resolution images.

Place recognition performance is assessed using the Recall@N score, which measures how often one of the top N images retrieved is within \SI{25}{\meter} of the query image location. In the case of temporal distance prediction, the top N images are those with the smallest predicted temporal distances.


\section{Results}
\label{sec:results}

\subsection{Robot navigation experiments}

Tables \ref{tab:indoors} and \ref{tab:outdoors} display the results of the navigation experiments with the Turtlebot and Summit XL.

\begin{table}[t]
    \centering
    \caption{\textbf{Indoors experiment} success rates over 60 repetitions, driven with the Turtlebot.}
    \label{tab:indoors}
    

\begin{tabular}{cccrccc}
    \toprule

    \multirow{2}{*}{Method} &
    \multirow{2}{*}{Type} &
    Temporal & & Easy & Hard & Total \\
    && filter &  \multicolumn{2}{l}{ $n=33$ } & $ 27 $ & 60 \\

  \midrule

  GNM~\cite{shahGNMGeneralNavigation2023} &
  \emph{T} &
  Window & &
  0.52 & 0.26 & 0.39 \\

  \midrule

  \multirow{2}{*}{\methodname} & 
  \multirow{2}{*}{\emph{P}} &
  Window & &
  0.62 & 0.60 & 0.61 \\

    &
   & Bayesian &&
   $\bm{0.65}$ & $\bm{0.77}$ & $\bm{0.69}$ \\

  \bottomrule

\end{tabular}


\end{table}

\begin{table}[t]
    \centering
    \caption{\textbf{Outdoors experiment} success rates over 60 repetitions, driven with the Summit XL.}
    \label{tab:outdoors}
    

\begin{tabular}{cccrccc}
    \toprule
   \multirow{2}{*}{Method} &
   \multirow{2}{*}{Type} &
   Temporal && Easy & Hard & Total \\
  && filter &  \multicolumn{2}{l}{ $n=24$ } & 36 & 60 \\

  \midrule

  GNM~\cite{shahGNMGeneralNavigation2023} & 
  \emph{T} &
  Window & &
  $\bm{0.67}$ & 0.33 & 0.47 \\

  \midrule

  \multirow{2}{*}{\methodname} &
  \multirow{2}{*}{P} &
  Window & &
  0.46 & 0.44 & 0.45 \\

   && Bayesian & &
   $\bm{0.67}$ & $\bm{0.53}$ & $\bm{0.58}$ \\
   
    \bottomrule
    \vspace{-0.2cm}
    \\
& \multicolumn{6}{r}{\emph{T: temporal distance, P: place recognition}}

\end{tabular}


\end{table}

\vspace{0.2cm}
\textbf{Indoors.}
The GNM baseline has a notanly lower success rate than the proposed place recognition based approaches.
\methodname~%
shows a 56\% SR increase, which rises to 77\% with the Bayesian filter. This observation is interesting given that GNM training includes indoor images from the GS4~\cite{hiroseDeepVisualMPCPolicy2019} dataset, while CosPlace models are trained on outdoor Google Streetview images. Place recognition models exhibit broad generalization capabilities, learning features that generalize across domains.

Similar to Shah~\etal~\cite{shahGNMGeneralNavigation2023}, we split the test routes into 'Easy' and 'Hard' categories in posterior analysis.
The categories are based on the number of narrow turns and tight passages along the routes. We chose a threshold that splits the routes into evenly sized groups. For indoor routes, this threshold was set to 4. 'Easy' routes had fewer than 4 such features, while 'Hard' routes had 4 or more.
The categorization provides further insight into the method performances. While the differences in SR are minimal for 'Easy' routes, the advantage of place recognition is evident in the 'Hard' category. GNM's SR decreases by 50\% from 'Easy' to 'Hard,' but \methodname~maintains the same SR, even improving when the Bayesian filter is employed. 

Introducing the Bayesian filter yields clear advantages over the sliding window method. Visual inspection confirms improved stability and performance during turns, especially in narrow gaps. The filter mitigates errors such as mid-turn direction changes that can arise because of the erratic behavior of the sliding window approach.

\begin{table*}[!t]
    \centering
    \caption{\textbf{ Offline evaluation} of Recall@1 for the place recognition (\emph{P}) and temporal distance (\emph{T}) models.}
    \label{tab:vpr}
    \begin{tabular}{lccccccccccccccc}

\toprule

&&&
\multicolumn{2}{c}{Multiview} & \multicolumn{7}{c}{Front-facing}
\\
\cmidrule(l{5pt}r{5pt}){4-5}
\cmidrule(l{5pt}r{5pt}){6-12}
\vspace{0.1cm}

\multirow{2}{*}{Method} &
\multirow{2}{*}{Type} &
\multirow{2}{*}{Resolution} &
\multirow{2}{*}{Pitts30k} &
Tokyo &
MSLS &
\multirow{2}{*}{St. Lucia} &
SVOX &
SVOX &
SVOX &
SVOX &
SVOX
\\
&&&& 24/7 & Val && Night & Overcast & Rainy & Snow & Sun
\\
\hline

CosPlace \cite{bertonRethinkingVisualGeolocalization2022a} &
\emph{P} &
full\textsuperscript{*} &

\textbf{89.5} &

\textbf{81.9} &

\textbf{80.8} &

\textbf{98.8} &

\textbf{32.9} &

\textbf{85.4} &

\textbf{79.6} &

\textbf{85.3} &

\textbf{61.5}
\\

\hline
CosPlace-LR &
\emph{P} &
$85\times64$ &

\underline{71.8} &

\underline{24.1} &

\underline{43.8} &

\underline{90.4} &

\underline{1.2} &

\underline{33.9} &

\underline{26.1} &

\underline{24.5} &

\underline{12.4}
\\

GNM \cite{shahGNMGeneralNavigation2023} &
\emph{T} &
$85\times64$ &

8.9 &

0.3 &

2.6 &

8.7 &

0.0 &

0.1 &

0.2 &

0.0 &

0.4
\\

\bottomrule
\vspace{0.1cm}
&&&&&&&&&
\multicolumn{3}{r}{\emph{*: various resolutions}}
\end{tabular}
\end{table*}

\vspace{0.2cm}
\textbf{Outdoors.}
Outdoors, the success rate gap between methods is narrower compared to indoors, despite CosPlace-LR's outdoor training data.
One possible explanation is the higher variation in waypoint estimation performance observed in the calibration tests. Consequently, the waypoint estimation module contributes more significantly to the final SR's, diminishing the effect of subgoal selection performance.
The magnitude of the performance increase brought by the Bayesian filter is consistent with the indoor experiments, the improvement being around 10 percentage points in both cases. 
The differences in SR's between the 'Easy' and 'Hard' categories are similar to the indoor experiments. GNM's SR drops by half from 'Easy' to 'Hard,' whereas \methodname~exhibits minimal to no performance decrease, emphasizing place recognition's effectiveness in maneuvers where having a correct subgoal is crucial.

\begin{figure}[ht]
  \centering
  
  \begin{adjustbox}{width=\linewidth, bgcolor=\mosaicbgcolor} 
    \begin{subfigure}{0.32\columnwidth} 
      \includegraphics[width=\linewidth]{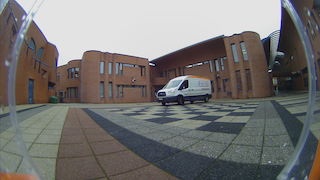} %
    \end{subfigure}
    \begin{subfigure}{0.32\columnwidth}
      \includegraphics[width=\linewidth]{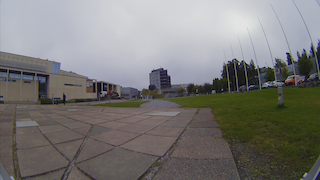} %
    \end{subfigure}
    \begin{subfigure}{0.32\columnwidth}
      \includegraphics[width=\linewidth]{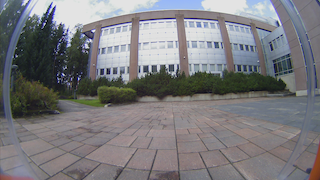} %
    \end{subfigure}
    \end{adjustbox}
  \caption{
  Repetitive patterns can disturb place recognition.}
  \label{fig:fail-routes}
\end{figure}

In the analysis of outdoor experiments, an even distribution of routes between the 'Easy' and 'Hard' categories was achieved by a threshold of 3 turns or narrow passages per route.
The occurrence of visually bursty content~\cite{jegouBurstinessVisualElements2009}, characterized by repetitive geometric patterns that heavily influence image embeddings, poses a challenge for place recognition methods~\cite[p.12]{masone_survey_2021} on certain test routes (see Fig.~\ref{fig:fail-routes}). This issue, causing the relatively low SR of
\methodname~%
with the sliding window in the 'Easy' category can lead to the sliding window method becoming trapped within the bursty map region, resulting in navigation failure. In contrast, the Bayesian filter handles bursty content more effectively by maintaining a full posterior belief across all map nodes and avoiding such entrapment. If the robot traverses the bursty segment successfully, the filter accurately localizes and completes the test route without issues.

\subsection{Offline evaluation}

Here, we present the results of evaluating GNM and CosPlace on the VPR-Benchmark. We also discuss the impact of input resolution and domain shift on recall rates.

\vspace{0.1cm}
\textbf{Performance Comparison.}
Table~\ref{tab:vpr} shows a comparison of the retrieval performances of the GNM and CosPlace models across several benchmark datasets. Notably, temporal distance prediction performs worse than place recognition across all datasets.
GNM's recall follows a similar trend as CosPlace-LR, with GNM achieving higher recall wherever CosPlace-LR excels. However, GNM's recall values are consistently an order of magnitude lower. For instance, on Tokyo24/7 and St. Lucia, where CosPlace-LR attains over 70\% recall, GNM only reaches approximately 9\%. On other datasets, GNM's performance is significantly lower.

\vspace{0.1cm}
\textbf{Impact of Input Resolution.}
Decreasing image resolution has a substantial impact on CosPlace's performance. Reducing the resolution to $85\times64$ pixels decreases recall rates up to 58 percentage points on datasets like Tokyo24/7, MSLS, and SVOX.
Interestingly, GNM's temporal distance prediction performs best on datasets where the performance differences between full and low-resolution CosPlace models are minimal, namely Pitts30k and St. Lucia. This suggests that GNM performance, too, would be improved by training the model with higher-resolution images.

\vspace{0.1cm}
\textbf{Viewpoint and Appearance Change.}
Pittsburgh30k and Tokyo24/7 datasets capture images from various angles, while the rest feature images from front-facing vehicle cameras. Despite the similarity in viewpoint variation between GNM's training data and front-facing datasets, this is not reflected in recall rates. GNM performs well on Pittsburgh30k but poorly on SVOX.
This discrepancy may stem from other factors contributing to domain shift between query and reference images. Besides viewpoint changes, Pittsburgh30k and St. Lucia exhibit limited variation. The other datasets contain shifts in illumination, weather, and camera which GNM struggles to handle, not having been explicitly trained for such invariance.

\begin{table}[!t]
    \centering
    \caption{\textbf{Runtimes} for the different methods with sliding window and Bayesian filter. {\scriptsize \emph{*: Bayesian filter considers all map nodes}}}
    \label{tab:runtimes}
    
    \resizebox{0.7\linewidth}{!}{

\begin{tabular}{cccc}
    \toprule

 \multirow{2}{*}{Method} & Temporal & Window &  Runtime \\
& filter & size &(\SI{}{\milli \second})\\

  \midrule

  \multirow{2}{*}{GNM~\cite{shahGNMGeneralNavigation2023}} & Window & 5 & 45 \\
   & Window & 21 & 174  \\

  \midrule

  \multirow{3}{*}{\methodname} & Window & 5 & \textbf{19} \\
    & Window & 21 & \textbf{19}  \\

   & Bayesian & ~-$^*$ &  \underline{41} \\
   
    \bottomrule
\end{tabular}
}
    
\end{table}

\subsection{Runtime analysis}

Table~\ref{tab:runtimes} shows average runtimes for
\methodname~and the baseline.
Replacing temporal distance prediction with place recognition significantly reduces runtime.
Place recognition's runtime does not depend on window size, while temporal distance is computed separately for each subgoal candidate inside the window.
%
%
The Bayesian filter increases \methodname's runtime slightly. This enables making a resource-performance trade-off based on the navigation requirements.

\section{Conclusion}
\label{sec:conclusion}
Our findings show that place recognition enables more accurate subgoal selection than the temporal distance prediction methods at a lower computational cost.
The offline evaluation implies that appearance change between the reference run and robot operation conditions would further amplify the difference.
These results suggest that the training of learning-based models for robotics should not be unnecessarily limited to robotics-specific data.
If some part of the robot's task can be cast as a more general learning problem, a larger amount of more diverse data can be used, leading to better performance.
Our future work will apply this principle to developing appearance-invariant subgoal selection models and goal-reaching policies.



\section*{ACKNOWLEDGMENT}

{\small
The authors thank Jani Käpylä, Olli Suominen, and Jussi Rantala from \href{https://civit.fi/}{CIVIT} for access to the Summit XL.
We would also like to thank María Andrea Cruz Blandón and German F. Torres for their valuable comments on an earlier version of the manuscript.
}


\def\url#1{}
\bibliographystyle{IEEEbst/IEEEtran}
\bibliography{references}

\end{document}